\documentclass[runningheads]{llncs}
\usepackage{comment}
\usepackage{graphicx}

\usepackage{amsfonts,amssymb}
\usepackage{amsmath}
\usepackage{mathtools}
\usepackage{subcaption}
\usepackage{marvosym}

\usepackage{hyperref}

\usepackage{multirow}
\usepackage{tabulary,booktabs}
\usepackage{caption} 
\makeatletter
\newif\if@restonecol
\makeatother

\usepackage[linesnumbered,ruled,vlined]{algorithm2e}
\usepackage{algpseudocode}
\usepackage{amsmath}

\let\OLDthebibliography\thebibliography
\renewcommand\thebibliography[1]{
	\OLDthebibliography{#1}
	\setlength{\parskip}{0pt}
	\setlength{\itemsep}{0pt plus 0.3ex}
}

\begin{document}
%
\title{Cross-domain Trajectory Prediction with CTP-Net}

\def\YOFOSubNumber{52}  

\titlerunning{CTP-Net}
%
\author{Pingxuan Huang\inst{1} \and
Zhenhua Cui\inst{2} \and
Jing Li\inst{3}\and
Shenghua Gao\inst{3}\and
bo Hu\inst{2}\and
Yanyan Fang\inst{2}\textsuperscript{,\Letter}
}
\authorrunning{Huang et al.}
%
\institute{University of Michigan, Ann Arbor, \email{pxuanh@umich.edu} \and
Fudan University, \email{\{19110720030,bohu,yyfang\}@fudan.edu.cn} \and
ShanghaiTech University, \email{\{lijing1,gaoshh\}@shanghaitech.edu.cn}}
\maketitle              

\begin{abstract}
Most pedestrian trajectory prediction methods rely on a huge amount of trajectories annotation, which is time-consuming and expensive. Moreover, a well-trained model may not effectively generalize to a new scenario captured by another camera. Therefore, it is desirable to adapt the model trained on an annotated source domain to the target domain. To achieve domain adaptation for trajectory prediction, we propose a Cross-domain Trajectory Prediction Network (CTP-Net). In this framework, encoders are used in both domains to encode the observed trajectories, then their features are aligned by a cross-domain feature discriminator.  
Further, considering the consistency between the observed and the predicted trajectories, a target domain offset discriminator is utilized to adversarially regularize the future trajectory predictions to be in line with the observed trajectories. Extensive experiments demonstrate the effectiveness of our method on domain adaptation for pedestrian trajectory prediction.

\end{abstract}
\section{Introduction}
\par
	Pedestrian trajectory prediction is attracting increasing attention these years due to its promising application prospects, such as autonomous driving~\cite{8793868}, safety monitoring \cite{doi:10.2514/6.2019-3514}, socially-aware robots \cite{luber2010people}, \emph{etc}. However, existing deep learning based trajectory prediction methods\cite{Yi2016Pedestrian}\cite{xu2018encoding}\cite{zhang2019sr}\cite{sadeghian2019sophie} usually need huge amounts of annotated data. When these methods are applied to a new scenario with different data distribution, the model trained on the original scenario may not generalize well and the performance degrades\cite{9527390}. Moreover, annotating the future trajectories for the new scenario is time-consuming and expensive. 
	Therefore, in this paper, we introduce a cross-domain trajectory prediction task, where both observed trajectories and annotated future trajectories are available in the existing scenario (\emph{i.e.}, source domain), but only the observed trajectories is available in the new scenario (\emph{i.e.}, target domain). Furthermore, the trajectories distribution varies between the source and the target domain. Accordingly, we aim to improve the model performance in the target domain, without any annotated future trajectory in the target domain. To the best of our knowledge, this is the first work to consider domain adaptation for trajectory prediction task.

	To perform domain adaptation in trajectory prediction, the Cross-domain Trajectory Prediction Network (CTP-Net) is proposed, which consists of two domain adaptation parts: the feature-level cross-domain alignment, and the target domain trajectory alignment. In detail, the feature-level cross-domain adaptation aims to align the features from the source domain and the target domain, while the target domain trajectory alignment is in charge of regularizing the consistency between the predicted future trajectories and the observed trajectories on the target domain.
	
	We first pre-train a source domain encoder-decoder to predict the future trajectories with supervision, under a standard training pipeline. Observed trajectories are encoded into a latent feature space, then the encoded feature is decoded to predict offsets of the future time-steps. Thereafter, the parameters of the source encoder-decoder are fixed in further training. The next step is to implement the feature-level cross-domain alignment by training a target encoder and a cross-domain feature discriminator. The discriminator aims to estimate the distribution distance between the source encoder features and the target encoder features. Then, the adversarial objective of the target encoder is to mitigate the distribution distance between source and target encoder features. Consequently, the target features are adversarially aligned with the source features.
	
	After training the target encoder, the target domain trajectory alignment is conducted by training a target decoder and a target domain offset discriminator. The purpose of the target domain offset discriminator is to estimate the distribution distance between the target domain observed trajectories offsets and the predicted offsets. As to the target decoder, it is composed of the fixed source decoder and a trainable coordinate offset adaptor aiming at transforming the source decoder prediction to the target domain. Especially, the offset adaptor is adversarially trained to mitigate the distribution distance between the predicted offsets and the target observed offsets. Therefore, the predicted future offsets are regularized to be consistent with that of the target domain observed trajectories.
	%
	%
	\par The main contributions of this paper are summarized as follows: i.) the domain adaptation is first introduced into trajectory prediction;  ii.) a Cross-domain Trajectory Prediction Network (CTP-Net) is proposed to adapt the model from the annotated source domain to the target domain by both feature-level cross-domain alignment and the target domain trajectory alignment; iii.) extensive experiments demonstrate the effectiveness of our method on domain adaptation for trajectory prediction.

\section{Related work}
\par  Recent works on deep-learning-based trajectory prediction and domain adaptation are introduced briefly.
	
\textbf{Trajectory prediction.} With the development of deep learning, many methods have been proposed in trajectory prediction area \cite{xu2018encoding}\cite{zhang2019sr}\cite{9156890}\cite{li2021temporal}\cite{tang2022evostgat}\cite{quan2021holistic}\cite{chen2022fully}. Early works \cite{8099976}\cite{yagi2018future} model person motions independently. Gradually, researchers \cite{alahi2016social}\cite{2016pedestrian} introduced the modeling of human-human interactions. Recently, there are a variety of perspectives to consider trajectory prediction, such as Imitative Decision Learning (IDL) ~\cite{li2019way}, Inverse Reinforcement Learning (IRL) \cite{9009551}, Social Interpretable Tree (SIT) \cite{DBLP:journals/corr/abs-2205-13296}, Constant Velocity method \cite{8972605}, and Graph-based methods \cite{zhou2021ast}\cite{li2022graph}, \emph{etc}.  Different from existing works, we introduce a cross-domain trajectory prediction network (CTP-Net), which utilizes feature-level cross-domain alignment and observation-future consistency to adapt the model from the labeled source domain to the unlabeled target domain.
	
%
	
\textbf{Domain adaptation.} Domain adaptation is one of the research focuses on Transfer Learning\cite{farahani2021brief}. Its main task is to enhance models’ performance on test (target) data by minimizing the marginal distribution difference between the training (source) data space and the test data space\cite{zhou2022domain}. Traditional methods, including the TCA \cite{DBLP:conf/ijcai/PanTKY09} and JDA \cite{6751384} tend to minimize a distribution distance metric. After the success of deep learning, neural networks are introduced to assist future extraction and automatic future adaptation. Many representative works, such as the DDC \cite{Tzeng2014deep}, and DAN \cite{pmlr-v37-long15} are proposed. Recently, with the development of Generative Adversarial Networks (GAN), such as the Wasserstein-GAN\cite{arjovsky2017wasserstein}, adversarial domain adaptation is becoming increasingly popular. Many classical structures, such as the DANN \cite{Muhammad10}, DSN \cite{Bousmalis10} have justified the rationality of their back stone idea: to minimize the difference between the source and target domain until the adversary domain discriminator could not distinguish them. Inspired by the achievement of domain adaptation, we proposed a task-oriented Cross-domain Trajectory Prediction Network (CTP-Net) to conduct trajectory prediction. To the best of our knowledge, our work is the first to successfully use domain adaptation in trajectory prediction.

\begin{figure}[t]
	\centering
	\includegraphics[scale=0.56]{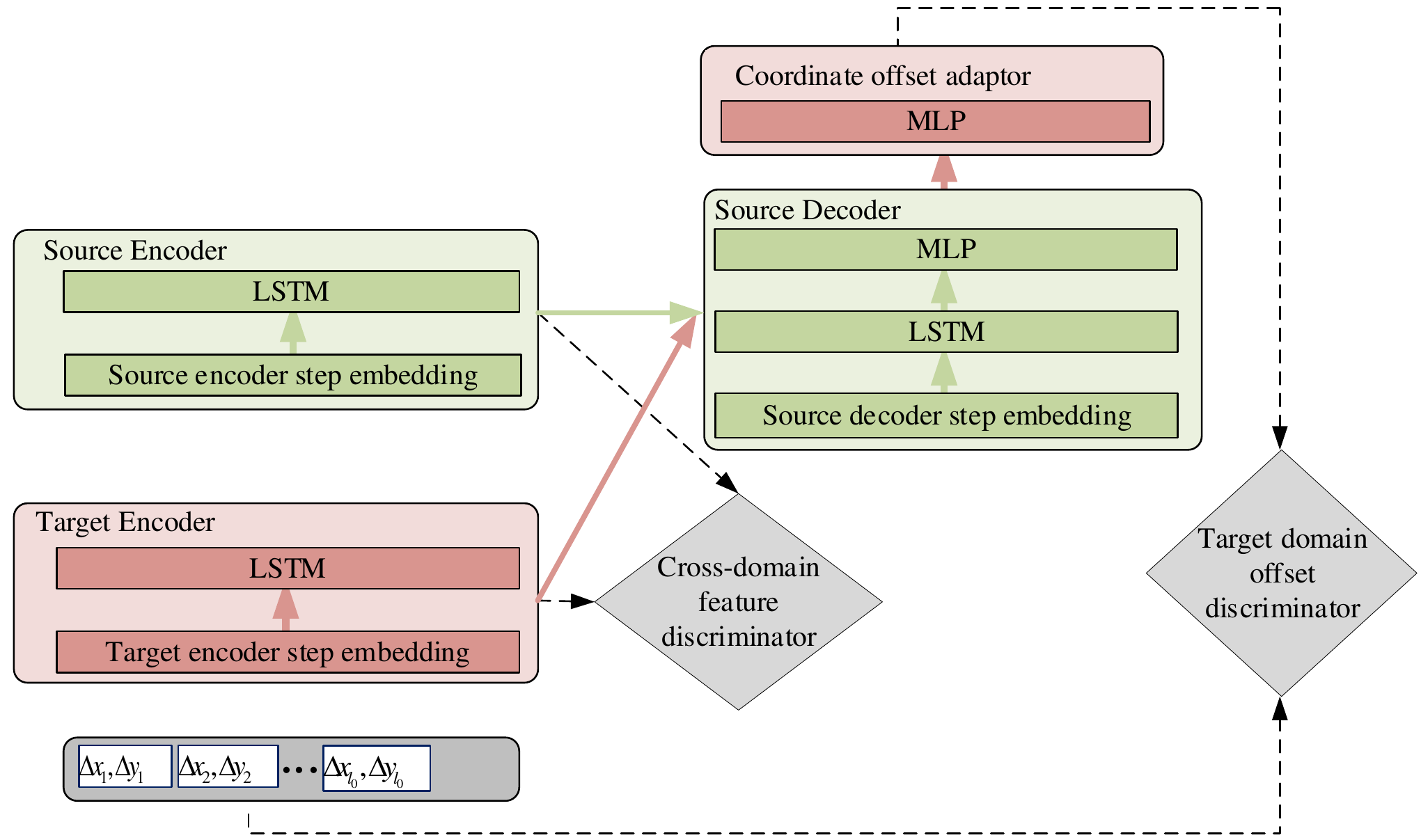}
	\caption{Structure of the CTP-Net module for Cross-domain Trajectory Prediction. Where \textit{green} represents source-related components; \textit{red} represents target-related components; \textit{gray} represents adaptation-related components; \textit{solid lines} represent data/feature flow for trajectory prediction; \textit{dashed lines} represent feature flow for domain adaptation.}
	\label{module}
\end{figure}

\section{Method}
	\subsection{Problem formulation}
	
	In our scenario, data comes from 2 domains: the Source domain($\mathbb{S}$), which has ground truth, and the Target domain($\mathbb{T}$), which is unlabeled. Under this setting, our goal is to predict the future trajectory of target domain data ($\mathbb{F}_t \in \mathbb{R}^{lf \times 2}$) with the target domain observation data ($\mathbb{O}_t \in \mathbb{R}^{lo \times 2}$).
	
	Previous trajectory prediction models are firstly trained on the source domain ($\mathbb{O}_s$ \& $\mathbb{F}_s$). Afterward, they will be directly applied to the target domain, which might miss the abundant information from the target domain. However, in our task, with the assistance of the encoder-decoder structure, as well as the domain adaptation method, we hope to construct a network that could successfully map the target data into the source trajectory feature domain ($\mathbb{M}_s$), then project the prediction trajectory back to the target domain. Specifically, instead of predicting absolute coordinate, we set the coordinate offset($\mathbb{C}$), which is the changing values between two consecutive coordinates, as our predicting stuff.

\subsection{Cross-domain Trajectory Prediction Network}
	
	The Cross-domain Trajectory Prediction Network (CTP-Net) contains two domain adaptation steps: the first is to align the cross-domain feature space, and the second is to align the coordinate offset space within the target domain. The whole pipeline of CTP-Net architecture is illustrated in Fig.\ref{module}. Roughly speaking, CTP-Net is composed of 3 components: 1. The source encoder-decoder; 2. The Cross-domain feature discriminator and the target encoder; 3. The coordinate offset  adaptor, and the target domain offset discriminator. 
	
	In this framework, we first pre-train the source encoder-decoder with a standard training pipeline on the source domain to predict the future trajectories. Afterward, we fix the source encoder-decoder parameters. Then we use the cross-domain feature discriminator and the target encoder to achieve feature-level cross-domain alignment. The purpose of this discriminator is to estimate the distribution distance between the source and the target feature domain. Finally, we adversarially train the target domain offset discriminator and the coordinate offset adaptor to align the coordinate offset within the target domain.

	\subsubsection {Wasserstein-distance}
	Similar to the KL-divergence, the Wasserstein-distance (W-distance) is a kind of metric to evaluate the difference between two distributions. First introduced by \cite{arjovsky2017wasserstein}, the W-distance has shown the advantages of adversary learning.
	The W-distance \cite{2017arXiv170701217S} between two distributions $\mathbb{P}_r$ and $\mathbb{P}_g$ could be calculated by:
	
	\begin{equation}
	W(\mathbb{P}_r, \mathbb{P}_g) = \frac{1}{K}\sum_{||D||_L\le K}\mathbb{E}_{x\sim\mathbb{P}_r}[D(x)] - \mathbb{E}_{x\sim\mathbb{P}_g}[D(x)]
	\label{2}
	\end{equation}
	
	Where $D(x)$ is a K-Lipschitz function, and it can be estimated by solving the following problem:
	\begin{equation}
	\max_{w:||D_w||_L\le K}\mathbb{E}_{x\sim\mathbb{P}_r}[D_w(x)] - \mathbb{E}_{x\sim\mathbb{P}_g}[D_w(x)]
	\end{equation}
	

	\subsubsection{Source Domain Training}
	
	This step is aimed to construct a trajectory prediction model on the source domain, which has an Encoder-Decoder structure.
	
	When it comes to our method, we construct both the source and the target encoder as the combination of a step embedding (SEB) and an LSTM ($k \in [1:lo]$), where $\theta_{se}$ is the source encoder parameter to be learned:
	\begin{equation}
	\begin{aligned}
	h_{sk},out_{sk} &= LSTM(h_{s(k-1)},SEB(o_{sk});\theta_{se})
	\end{aligned}
	\end{equation}
	
	The final hidden state of the encoder i.e. $h_{slo}$, is defined as the trajectory feature($m_{s}$). The decoder part is composed of a step embedding, an LSTM, and an MLP($\psi$). Specially, both the input and output of decoder are coordinate offsets ($\delta o_s/\delta f_s$, $k \in [lo+1:lo+lf]$):
	
	\begin{equation}
	\begin{aligned}
	h_{sk},out_{sk} = LSTM&(h_{s(k-1)},SEB(\delta o_{s(k-1)}/ \delta f_{s(k-1)});\theta_{sd})\\
	\delta f_{s(k)} &= \psi(out_{sk};\theta_{sd})
	\end{aligned}
	\end{equation}

\subsubsection{Trajectory Feature Domain Adaptation}
	\label{sec:tedca}
	
	The purpose of this step is to map the trajectory feature from the target domain into the source domain. Therefore, we adapt it to a Domain Adaptation task, which means the task of this step is to minimize the data distribution difference between $\mathbb{M}_s$ and $\mathbb{M}_t$. Accordingly, with the target decoder, we could estimate $Pr(m_s|o_t)$.
	
	To achieve this goal, we employ the W-distance-based Adversarial Domain Adaptation. Specifically, this step needs to train the target encoder with parameter $\theta_{te}$, and the cross-domain feature discriminator with parameter $\theta_{dA}$.
	According to  Eq. (\ref{2}): denoting the data distribution on $\mathbb{M}_s$ and $\mathbb{M}_t$ as $\mathbb{P}_{ms}$ and  $\mathbb{P}_{mt}$, the W-distance between $\mathbb{P}_{ms}$ and $\mathbb{P}_{mt}$ could be estimated by:

    \begin{equation}
	W(\mathbb{P}_{ms},\mathbb{P}_{mt}) \approx \mathbb{E}_{x\sim\mathbb{P}_{ms}}[\hat{D}_w(x)] - \mathbb{E}_{x\sim\mathbb{P}_{mt}}[\hat{D}_w(x)]
	\end{equation}
	
	where $\hat{D}_w(x)$, could be estimated by solving the problem:
	\begin{equation}
	\min_{w:||D_w||_L\le 1} L_{wd} =  \mathbb{E}_{x\sim\mathbb{P}_{mt}}[D_w(x)] - \mathbb{E}_{x\sim\mathbb{P}_{ms}}[D_w(x)]
	\end{equation}

     To enforce the constraint of the Lipschitz Function, the Gradient Penalty\cite{wgan_gp} is used, and the following problem is obtained:
	\begin{equation}
	L_{gp}=\mathbb{E}_{\hat{x}\sim\mathbb{P}_{\hat{x}}}[(||\nabla_{\hat{x}}D(\hat{x})||_2-1)^2]
	\end{equation}
	$\mathbb{P}_{\hat{x}}$ is uniformly distributed on the straight line between pairs of points sampled from $\mathbb{P}_{ms}$ and $\mathbb{P}_{mt}$. Eventually, we should solve the problem:
	
	\begin{equation}
	\min_{D(x;\theta_{dA})}L_{wd} - \lambda L_{gp}
	\end{equation}
	After getting the estimation of the distribution distance, the task of the target encoder is to minimize the distribution difference between $\mathbb{M}_s$ and $\mathbb{M}_t$, i.e.
	
	\begin{equation}
	\min_{\theta_{te}}(-\mathbb{E}_{x\sim\mathbb{P}_{o_t}}[\hat{D}(T E_{\theta_{te}}(x))])
	\end{equation}

Where $\mathbb{P}_{o_t}$ is the data distribution of $\mathbb{O}_t$, \textit{TE} is target encoder. The detailed algorithm to train the cross-domain feature discriminator ($D_{\theta_{dA}}$), and the target encoder is shown as \hyperref[alg:tedca]{Algorithm 1}. In our network, the structure of the target encoder is the same as that of the source encoder, and the input of the Cross-domain feature discriminator is the concatenation of the $m_s$/$m_t$.
	
	\begin{algorithm}[ht]
		\caption{CTP-Net cross-domain feature alignment algorithm}
		\label{alg:tedca}
		\KwIn{The learning rate of cross-domain feature discriminator, and target encoder: $\alpha$, $\beta$; The gradient penalty coefficient $\lambda$; batch size $m$; adversarial training epoch $N$, domain critic training iteration $n$.}
		Fix $\theta_{se}$;\\
		\For{T = 1,2 \dots N }
		{
			\For{t = 1,2 \dots n}
			{
				$L\leftarrow 0$;\\
				\For{i = 1,2\dots m}
				{
					Sample $o_s$ from $\mathbb{O}_s$, sample $o_t$ from $\mathbb{O}_t$, generate random number $\gamma \sim U[0,1]$;\\
					$m_s$ $\leftarrow$ Source Encoder($o_s$);\\
					$m_t$ $\leftarrow$ Target Encoder($o_t$);\\
					$m_m$ $\leftarrow$ $\gamma m_s + (1-\gamma) m_t$;\\
					$L\leftarrow L + D_{\theta_{dA}}(m_t) - D_{\theta_{dA}}(m_s) + \lambda(||\nabla_{m_m} D_{\theta_{dA}}(m_m)||_2-1)^2$;\\
				}
				$\theta_{dA} \leftarrow RMSprop(\nabla_{\theta_{dA}}\frac{1}{m}L;\alpha)$;
			}
			Sample $\{o_t\}_{i=1}^m$ from $\mathbb{O}_t$;\\
			$\theta_{te} \leftarrow RMSprop(\nabla_{\theta_{te}}\frac{1}{m}\sum_{k=1}^m-D_{\theta_{dA}}($Target Encoder$(o_t));\theta_{te},\beta)$;
		}
	\end{algorithm}

	\subsubsection{Coordinate offset Domain Adaptation}
	
	The purpose of the Coordinate offset Adaptor (CA) is to project the coordinate offset from the source domain to the target domain, which is a part of Domain Adaptation, and a similar structure of \hyperref[sec:tedca]{the previous part} is employed. Instead of the absolute pedestrian coordinate, we select the coordinate offset as the input and the output of both the decoder and the CA. Both the theoretical and the experimental analysis demonstrate the rationality of this setting. To be specific, in most circumstances, it is reasonable to assume that the distribution of the coordinate offset is a symmetric distribution with mean zero. Because pedestrians could go both up/down, and right/left freely. On the contrary, the variation of starting coordinates will significantly enlarge the sample space and complicate the absolute coordinate distribution. Accordingly, coordinate offset space is more suitable for domain adaptation tasks. Furthermore, the ablation experiment also justifies this setting.
	
	Since the $lo$ is usually different from $lf$ (in our experiment, $lo = 8$, $lf=12$), given the requirement of training the target domain offset discriminator, the input frame length of CA might not be 12. In our model, we set the CA input as the concatenation of 6 consecutive frames.

\subsubsection{Inference}
	
	Previous sections show the details to construct a CTP-Net that could map the trajectory feature from the target domain to the source domain, as well as project the coordinate offset from the source domain to the target domain. When finishing training, the inference procedure on the target domain is: Target Encoder $\rightarrow$ Source Decoder $\rightarrow$ Coordinate Offset Adaptor.

	\section{Experiments}
	
	\subsection{Datasets and Evaluation Metrics}
	
\begin{figure}[ht]
    \begin{subfigure}{.48\textwidth}
      \centering
      \includegraphics[width=6cm]{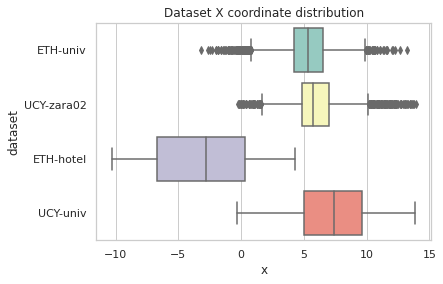}  
      \caption{X coordinates boxplot}
      \label{fig:coor_dist_sub_x}
    \end{subfigure}
    \quad
    \begin{subfigure}{.5\textwidth}
      \centering
      \includegraphics[width=6cm]{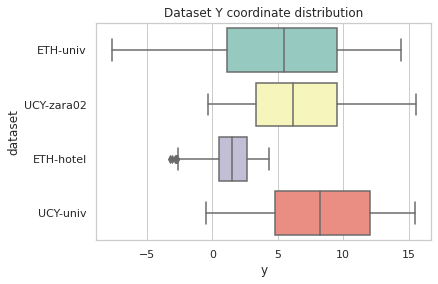}  
      \caption{Y coordinates boxplot}
      \label{fig:coor_dist_sub_y}
    \end{subfigure}
    \caption{Trajectories coordinates distribution}
    \label{fig:coor_dist}
    \end{figure}

	\textbf{Datasets:} We evaluate proposed model on the following public datasets: \textbf{ETH} \cite{pellegrini2009you} and \textbf{UCY} \cite{lerner2007crowds}. The ETH dataset contains two scenes named ETH-univ and ETH-hotel, while the UCY dataset contains three scenes named UCY-zara01, UCY-zara02, and UCY-univ. Especially, we conduct experiments on the following 4 datasets: ETH-univ, ETH-hotel, UCY-zara02, and UCY-univ. Figure \ref{fig:coor_dist} presents the X/Y coordinates distribution for each dataset.
	
    
	\noindent{\textbf{Evaluation metrics:}} Following previous works \cite{pellegrini2009you}\cite{alahi2016social}, we adopt two common metrics for testing. (1) \emph{ Average Displacement Error (ADE)}: is the mean squared error (MSE) between the ground truth and the prediction over all estimated time steps. (2)  \emph{Final Displacement Error (FDE)}: is the Euclidean distance between the prediction and the ground truth at the final prediction. They are defined as:
	\begin{equation}
	\begin{aligned}
	ADE &= \frac{\sum_{i=1}^N\sum_{t=lo}^{lo+lf-1} || \hat{F}_{t+1}^i-F_{t+1}^i||_2}{N*lf} \\
	FDE &= \frac{\sum_{i=1}^N||\hat{F}_{lo+lf}^i-F_{lo+lf}^i ||_2}{N}
	\end{aligned}
	\end{equation}
	
	Where $N$ is the sample size, $lf$ is the number of predicted frames, $lo$ is the number of observed frames, $\hat{F}_{t+1}^i$ is the predicted trajectory coordinates of the $i^{th}$ pedestrian at time $t+1$, and $F_{t+1}^i$ is the corresponding ground truth.
	
	\subsection{Experimental Setup}
	According to Figure \ref{fig:coor_dist}, the coordinates distribution of ETH-univ is similar to that of UCY-zara02, but the coordinates distribution of ETH-hotel is much different from that of UCY-univ. Therefore, we construct 4 groups of experiments, so that to test our method under \textbf{both the similar and dissimilar} distribution. \\
	1. Source: UCY-zara02; Target: ETH-univ.\\
	2. Source:  ETH-univ;    Target: UCY-zara02.\\
	3. Source: UCY-univ;   Target: ETH-hotel.\\
	4. Source: ETH-hotel;  Target: UCY-univ.
	
	For each source domain dataset, we separate the $training: validation$ as $8:2$;
	For each target domain dataset, we separate the $training: test$ as $4:6$.
	Therefore, models are trained on the source training and the target training set, validated on the validation set, and tested on the test set. Following the common setting \cite{alahi2016social}, we set the observation frame length as 8 (3.2 seconds), and the prediction frame length as 12 (4.8 seconds).
	\vspace{-4pt}		
	\subsection{Implementation Details}
	In our experiments, we set the step embedding size as 32, the LSTM hidden size as 512, and the layer amount of the source decoder MLP as 3. Moreover, all the offset adaptor, the target domain offset discriminator, and the cross-domain feature discriminator are constructed by MLP with ReLU functions, and the layer amounts for them are 2, 10, and 5.

	\begin{table}[t]
    \begin{tabular}{l|cc|cc|cc|cc|cc|cc|cc|cc}
        \hline
        \multirow{3}{*}{} & \multicolumn{8}{c|}{ADE}                                                                                                                                 & \multicolumn{8}{c}{FDE}                                                                                                                                   \\ \cline{2-17} 
                   & \multicolumn{2}{c|}{eth-zara}         & \multicolumn{2}{c|}{zara-eth}        & \multicolumn{2}{c|}{stu-hotel}        & \multicolumn{2}{c|}{hotel-stu}       & \multicolumn{2}{c|}{eth-zara}         & \multicolumn{2}{c|}{zara-eth}         & \multicolumn{2}{c|}{stu-hotel}        & \multicolumn{2}{c}{hotel-stu}        \\
        \multicolumn{1}{c|}{}                       & Val                  & Test          & Val                  & Test         & Val                  & Test          & Val                  & Test          & Val                  & Test          & Val                  & Test          & Val                  & Test          & Val                  & Test          \\
        \hline
        SO                                          & 1.73                 & 2.76          & 0.88                 & 2.90          & 0.81                 & 2.76          & 0.74                 & 4.63          & 3.06                 & 4.89          & 1.46                 & 5.28          & 1.55                 & 5.14          & 1.26                 & 8.46          \\
        F-T                                         & \multicolumn{1}{l}{} & 3.61          & \multicolumn{1}{l}{} & 2.98         & \multicolumn{1}{l}{} & 2.56          & \multicolumn{1}{l}{} & 2.50          & \multicolumn{1}{l}{} & 5.55          & \multicolumn{1}{l}{} & 5.08          & \multicolumn{1}{l}{} & 5.11          & \multicolumn{1}{l}{} & 3.85          \\
        S-L                                         & 4.01                 & \textbf{1.27} & 0.92                 & 3.65         & 1.17                 & 2.17          & 1.00                    & 2.01          & 6.32                 & \textbf{1.83} & 1.47                 & 6.10           & 2.07                 & 3.86          & 1.52                 & 3.30           \\
        \hline
        TE                                          & \multicolumn{1}{l}{} & 3.34          & \multicolumn{1}{l}{} & 2.60          & \multicolumn{1}{l}{} & 2.79          & \multicolumn{1}{l}{} & 3.89          & \multicolumn{1}{l}{} & 5.65          & \multicolumn{1}{l}{} & 4.85          & \multicolumn{1}{l}{} & 5.19          & \multicolumn{1}{l}{} & 7.07          \\
        TO                                          & \multicolumn{1}{l}{} & 1.42          & \multicolumn{1}{l}{} & \textbf{2.40} & \multicolumn{1}{l}{} & \textbf{1.59} & \multicolumn{1}{l}{} & \textbf{1.61} & \multicolumn{1}{l}{} & 2.73          & \multicolumn{1}{l}{} & \textbf{4.35} & \multicolumn{1}{l}{} & \textbf{2.79} & \multicolumn{1}{l}{} & \textbf{2.93}\\
        \hline
    \end{tabular}
    \caption{Comparative experiment results between our model and baselines. \textbf{SO}: \emph{Source Only}, source encoder-decoder; \textbf{F-T}: \emph{Fine-Tuning}, fine-tuning based on trained source encoder-decoder; \textbf{S-L}: \emph{Social-LSTM}; \textbf{TE}: \emph{Target Encoder}, replace source encoder with target encoder;  \textbf{TO}: \emph{Target encoder + Offset adaptor}, final CTP-Net inference path.}
	\label{baseline}
\end{table}
	
	\subsection{Results}
	We compared the performance with 2 models, the results\footnote{For the convenience of reading, we refer to the dataset with its corresponding abbreviation: \textit{eth}: ETH-univ; \textit{hotel}: ETH-hotel; \textit{zara}: UCY-zara02; \textit{stu}: UCY-univ.} are shown in Table\ref{baseline}:
	\begin{itemize}
	    \item Social-LSTM \cite{alahi2016social}: With the assistance of pooling layer, the Social-LSTM vectorizes and utilizes the interaction between pedestrians.
	    \item Fine-tuning: This model is fine-tuned on the target training set. Specifically, based on the trained source encoder-decoder, this model takes the first 4 frames from the target observation as input and predicts the next 4 frames.
	\end{itemize}
	
	For both the Source-Only model and the Social-LSTM, in most comparison settings, there is a significant performance gap between Test (target) and Validation (source) set, which justifies the necessity of domain adaptation. Moreover, our schema is able to outperform the baselines in most cases, which is especially true when the source and target distributions are dissimilar to each other.

	\subsection{Ablation $\&$ Visulazation}

	\begin{table}[t]
	\centering
	\setlength{\tabcolsep}{1pt}
    \begin{tabulary}{\textwidth}{l|cc|cc|cc|cc|cc|cc|cc|cc}
	\hline
    \multicolumn{1}{c|}{\multirow{3}{*}{}} & \multicolumn{8}{c|}{ADE}                                                                                                     & \multicolumn{8}{c}{FDE}                                                                                                     \\
    \hline
    \multicolumn{1}{c}{}                       & \multicolumn{2}{|c|}{eth-zara} & \multicolumn{2}{c|}{zara-eth} & \multicolumn{2}{c|}{stu-hotel} & \multicolumn{2}{c|}{hotel-stu} & \multicolumn{2}{c|}{eth-zara} & \multicolumn{2}{c|}{zara-eth} & \multicolumn{2}{c|}{stu-hotel} & \multicolumn{2}{c}{hotel-stu} \\
    \multicolumn{1}{c|}{}                       & CO       & OF                & CO       & OF                & CO        & OF                & CO        & OF                & CO       & OF                & CO                & OF       & CO        & OF                & CO        & OF                \\
    \hline
    val                                        & 2.32     & 1.73              & 1.18     & 0.88              & 0.79      & 0.81              & 1.20      & 0.74              & 2.90     & 3.06              & 1.36              & 1.46     & 1.41      & 1.55              & 1.51      & 1.26              \\
    \hline
    SO                                         & 3.90     & 2.76              & 3.24     & 2.90              & 3.90      & 2.76              & 10.39     & 4.63              & 5.65     & 4.89              & 4.63              & 5.28     & 4.72      & 5.13              & 10.37     & 8.46              \\
    TE                                         & 4.12     & 3.34              & 3.36     & 2.61              & 4.01      & 2.79              & 11.02     & 3.89              & 5.70     & 5.65              & 4.45              & 4.85     & 4.44      & 5.19              & 12.48     & 7.06              \\
    TO                                         & 4.52     & \textbf{1.42}     & 3.68     & \textbf{2.40}     & 3.47      & \textbf{1.59}     & 4.66      & \textbf{1.61}     & 4.83     & \textbf{2.73}     & \textbf{4.32}     & 4.35     & 3.90      & \textbf{2.79}     & 5.55      & \textbf{2.93}\\
    \hline
    \end{tabulary}
	\caption{Ablation experiment results. \textbf{SO}: \emph{Source Only}; \textbf{TE}: \emph{Target Encoder};  \textbf{TO}: \emph{Target encoder + Offset adaptor}; \textbf{CO}: \emph{COordinate}; \textbf{OF}: \emph{OFfsets}.} 
	\label{ablation}
\end{table}
    
    \begin{figure}[t]
    \begin{subfigure}{.47\textwidth}
      \centering
      \includegraphics[width=5.5cm]{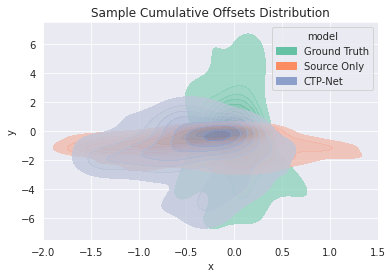}  
      \caption{Target sample cumulative offsets distributions of the ground truth, source-only, and CTP-Net prediction.}
      \label{fig:target_offsets_sub1}
    \end{subfigure}
    \quad
    \begin{subfigure}{.47\textwidth}
      \centering
      \includegraphics[width=5.5cm]{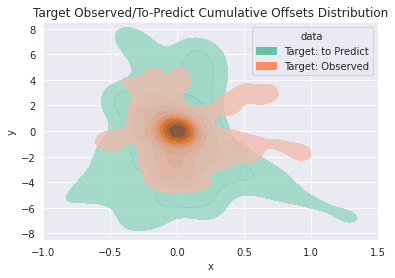}  
      \caption{Target-train observation and to-predict data cumulative offsets distributions}
      \label{fig:target_offsets_sub2}
    \end{subfigure}
    \caption{Cumulative offsets distributions}
    \label{fig:target_offsets}
    \end{figure}

	\par In the ablation experiments, we examined the efficacy of the 2 adaptation stages and compared the performance of using offset or coordinate as the prediction objective, results are shown in Table\ref{ablation}. Although 2-stage adaptation  (TO) significantly boosts performance, only the first adaptation stage (TE) has little or even negative effect. Moreover, the result demonstrates the coordinate offset is more appropriate than the absolute coordinate for adaptation.
	
	\par To intuitively verify the efficacy of CTP-Net, we visualized the distribution of target sample cumulative offsets in Figure \ref{fig:target_offsets_sub1}. To be specific, we subtract each step of predicted trajectories/ground truth by their last observation step, which is equal to moving the start prediction point to $(0,0)$, then we draw the \textit{Kernel Distribution Estimate-plot}\footnote{\url{https://seaborn.pydata.org/generated/seaborn.kdeplot.html}} for each model.	Based on the visualization result, compared with the Source-Only model, after adaptation, our prediction distribution is much closer to the target ground truth, which is especially true in the region of $y\le0$ and $x\ge0$. To further understand the adaptation result, we additionally visualized the cumulative offsets distribution of the target-train to-predict and observation data, which is used in the second adaptation stage. The result is shown in Figure \ref{fig:target_offsets_sub2}, which is equal to moving the start observation point of each trajectory to $(0,0)$ and then drawing their distribution plots. We figured out that the observation cumulative offsets distribution is closer (has more overlap) with the source-only cumulative offsets distribution in the $y\le0$ and $x\ge0$ region. Since the adaptation is with a higher probability of success \cite{analysis_of_representation_for_da} when the two regions are similar, the CTP-Net has the potential to achieve better performance in these 2 areas.

	
	\section{Conclusion}
	In this work, we introduce domain adaptation task into pedestrian trajectory prediction, where the source domain is fully annotated, but the target domain is unlabeled. In order to adapt the model from the source domain to the target domain, Cross-domain Trajectory Prediction Network (CTP-Net) is proposed. In the CTP-Net, two parts of adaptations are conducted: To align the target encoder features with the source encoder features distribution, the feature-level cross-domain alignment is performed; To make the predicted future trajectories consistent with the observed trajectories on the target domain, the target domain trajectory alignment is further implemented in the CTP-Net. Experiments justify the CTP-Net on domain adaptation for pedestrian trajectory prediction.

%
%
%
%

\bibliographystyle{splncs04}
\bibliography{bibliography}

\end{document}